\pdfoutput=1

\documentclass[11pt]{article}

\usepackage{acl}

\usepackage{times}
\usepackage{latexsym}
\usepackage{graphicx}
\usepackage{colortbl}
\usepackage{multirow}
\usepackage{booktabs}
\usepackage{subcaption}
\usepackage{rotating}
\usepackage[export]{adjustbox}

\usepackage{algorithm}
\usepackage{algpseudocode}
\usepackage{float}
\usepackage{pdflscape}
\usepackage{hyperref}
\usepackage[]{acronym}
\usepackage{amsmath}

\usepackage{amssymb}
\newtheorem{prop}{Proposition}

\newtheorem{proof}{Proof}

\usepackage{comment}
\usepackage[T1]{fontenc}

\usepackage[utf8]{inputenc}

\usepackage{microtype}

\usepackage{inconsolata}

\usepackage{graphicx}

\title{Adaptive Contrastive Search: Uncertainty-Guided Decoding for Open-Ended Text Generation}

\author{
  \textbf{Esteban Garces Arias\textsuperscript{1,2}},
  \textbf{Julian Rodemann\textsuperscript{1}},
  \textbf{Meimingwei Li\textsuperscript{1}},
  \textbf{Christian Heumann\textsuperscript{1}},\\
  \textbf{Matthias Aßenmacher\textsuperscript{1,2}}
\\
\\
  \textsuperscript{1}Department of Statistics, LMU Munich,
  \textsuperscript{2}Munich Center for Machine Learning (MCML)\\
\\
  \small{
    \textbf{Correspondence:} \href{mailto:esteban.garcesarias@stat.uni-muenchen.de}{esteban.garcesarias@stat.uni-muenchen.de}
  }
}

\begin{document}
\maketitle
\begin{abstract}

Despite the remarkable capabilities of large language models, generating high-quality text remains a challenging task. Numerous decoding strategies—such as beam search, sampling with temperature, top‑$k$ sampling, nucleus (top‑$p$) sampling, typical decoding, contrastive decoding, and contrastive search—have been proposed to address these challenges by improving coherence, diversity, and resemblance to human-generated text. In this study, we introduce Adaptive Contrastive Search (ACS), a novel decoding strategy that extends contrastive search (CS) by incorporating an adaptive degeneration penalty informed by the model's estimated uncertainty at each generation step. ACS aims to enhance creativity and diversity while maintaining coherence to produce high-quality outputs. Extensive experiments across various model architectures, languages, and datasets demonstrate that our approach improves both creativity and coherence, underscoring its effectiveness in text-generation tasks. We release our code, datasets, and models to facilitate further research.\end{abstract}

\section{Introduction}
\label{sec:introduction}

The Transformer \citep{vaswani2017attention} plays a key role in various generative natural language processing (NLP) tasks, such as generating stories, completing contextual text, and dialogue systems. However, the conventional method of training these models using maximum likelihood estimation (MLE) and decoding to the most probable sequence often results in substantial shortcomings. This can lead to repetitive and uncreative outputs, also known as \textit{degenerate} text.

\begin{figure}[ht]
\centering
\includegraphics[width = \linewidth, keepaspectratio]
{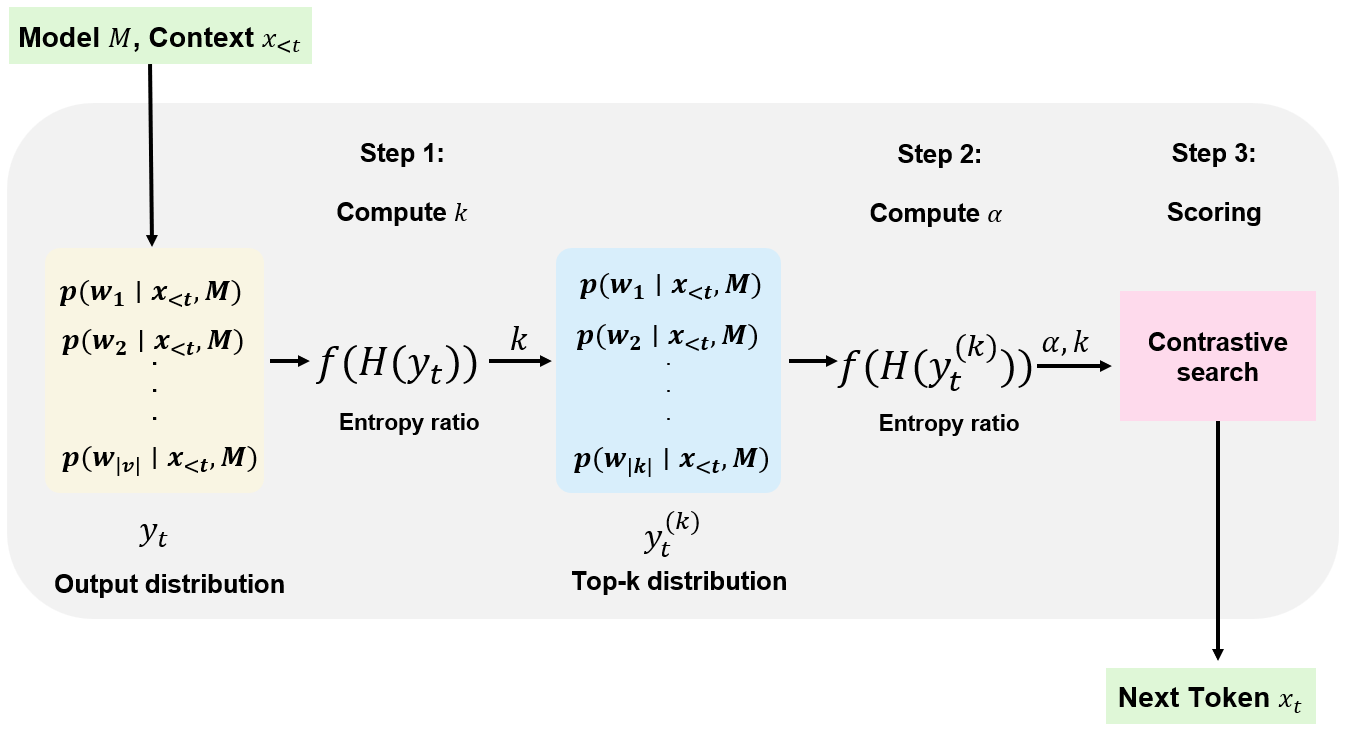}
\label{fig:figure_one}
\caption{Visualization of the Adaptive Contrastive Search (ACS) process: A three-step procedure that uses entropy as a proxy for model uncertainty to automatically adjust contrastive search parameters.}
\end{figure}

To address this issue, previous approaches have aimed to adjust the decoding strategy by incorporating sampling from less probable vocabularies. While this helps reduce repetitiveness, it introduces the problem of semantic inconsistency \citep{welleck2020consistency}. The sampled text may stray from or even contradict the original context provided by a human-written prompt.

In response, contrastive search \citep[CS,][]{su2022contrastive} has been introduced. It employs a fixed balance between model confidence and degeneration penalty throughout the generation process, maintaining a blend of likelihood and diversity.
However, it is important to note that this fixed weighting requires hyperparameter tuning and overlooks the unique demands of each generation step, where a different balance between model confidence and degeneration penalty might be desirable and even more effective.

We address this limitation by proposing adaptive contrastive search (ACS), an adaptive approach that automatically adjusts the hyperparameters of conventional CS. This method evaluates the model's uncertainty at each generation step and adjusts the weighting of both components without the need for manual intervention. The experimental outcomes demonstrate the effectiveness of our method, as it performs well in the task of open-ended text generation across different architectures, languages, and datasets.

\paragraph{Contributions} Our contributions can be summarized as follows:

\noindent \textbf{1.} We introduce an adaptive CS method based on the work by \citet{su2022contrastive} that measures the uncertainty of the model at each time step to automatically adjust the number of candidate tokens and the degeneration penalty.

\noindent \textbf{2.}  We conduct comprehensive experiments to compare our approach to various established decoding
methods, such as nucleus sampling \citep{holtzman2019curious}, contrastive decoding \citep[CD,][]{li2023contrastive}, and CS \citep{su2022contrastive}, for open-ended text generation.

\noindent \textbf{3.}  We offer new insights into MAUVE and its correlation with human judgments, highlighting the need for a more robust metric that better aligns with human preferences when evaluating decoding strategies for open-ended text generation. 

\noindent \textbf{4.}  Our code and datasets and results are publicly available under \href{https://github.com/YecanLee/Adaptive-Contrastive-Search}{this link}.

\section{Related work}
\label{sec:related_work}

Decoding methods are generally categorized into two types: deterministic and stochastic.

\paragraph{Deterministic Methods.} These approaches focus on choosing the text continuation with the highest probability according to the model's probability distribution. Prominent examples are beam search and greedy search. Recently, studies by \citet{shao2017generating,vijayakumar2018diverse,paulus2017deep,klein2017opennmt} have demonstrated that solely maximizing the output probability frequently leads to degenerated or repetitive text sequences, a problem that has been addressed by stochastic, sampling-based methods.

\paragraph{Stochastic Methods.} Top-\( k \), proposed by \citet{fan2018hierarchical}, samples from a subset of tokens \( V^{(k)} \) that represent the tokens with the higher scores in the output distribution.  Alternatively, nucleus-$p$ sampling \citep{holtzman2019curious} samples from the smallest subset \( S \) with a total probability mass above a threshold $p$; specifically, \( S \) is the smallest subset such that the cumulative probability for tokens in $S$ surpasses $p$. While these methods reduce model degeneration, the inherent stochasticity can lead to semantic divergence or disconnection from the human-written prompt.

To tackle the imbalance between coherence and diversity, methods such as typical sampling \citep{meister2023locally} and CD have been developed to produce more diverse and interesting text in open-ended settings. Typical sampling aims at generating based on the information content, which should be close to the expected information content, i.e., the conditional entropy of the model. CD, on the other hand, employs an expert language model (LM) and an amateur LM in parallel and searches for text that maximizes the difference between the expert's and the amateur's log probabilities, subject to plausibility constraints.

Our study, however, focuses on the work of \citet{su2022contrastive}, where they introduce Contrastive Search. In CS, given the prompt text $\boldsymbol{x}_{<t}$, the selection of the output token $x_t$ follows:
\vspace{-.3cm}
\begin{align}
x_t = \underset{v \in V^{(k)}}{\arg \max} \Bigg\{(1-\alpha) \times \underbrace{p_\theta(v \mid \boldsymbol{x}_{<t})}_{\text{model confidence}} -
\notag
\\
\alpha \times \underbrace{\left(\max \{s(h_v, h_{x_j}): 1 \leq j \leq t-1\}\right)}_{\text{degeneration penalty}} \Bigg\}
\label{eq:cs}
\end{align}

where $V^{(k)}$ is the set of top-$k$ predictions from the LM's probability distribution $p_\theta\left(\cdot \mid \boldsymbol{x}_{<t}\right)$. In Eq. \eqref{eq:cs}, the first term, model confidence, is the probability of the candidate $v$ predicted by the LM. The second term, degeneration penalty, measures how discriminative is the candidate $v$ with respect to the previous context $\boldsymbol{x}_{<t}$ and $s(\cdot, \cdot)$ computes the cosine similarity between token representations. Intuitively, a larger degeneration penalty of $v$ means it is more similar to the context, therefore more likely leading to undesirable repetitions in the generated output. The hyperparameter $k \in \mathbb N_{> 0}$ determines the number of candidate tokens to be considered, while $\alpha \in[0,1]$ regulates the importance of these two components.

Empirical results suggest different values for $\alpha$ and $k$, depending on the task, the datasets, and the language of interest, respectively \citep{su2022contrastive,su2022empirical,su2023contrastive}. An empirical study comparing CS and CD \cite{su2023contrastive} highlights the strengths and weaknesses of both approaches. The automatic evaluation results suggest that CD performs better on MAUVE \citet{pillutla2021mauve}, while CS excels in diversity and coherence. Additionally, through extensive human evaluations, they demonstrate that human annotators universally prefer CS over CD by substantial margins. Given the contradictory results between MAUVE and human evaluations, their analysis reveals that balancing diversity and coherence metrics better correlates with human judgments.

Further studies have extended the concept of CS to incorporate additional criteria in scoring candidates for the next token. \citet{chen2023fidelity} investigate the effect of adding a third criterion, fidelity, beyond model confidence and degeneration penalty to enhance the coherence of the generated text. This third criterion is again weighed by a hyperparameter \(\beta\) that is determined from empirical results. To the best of our knowledge, adaptive approaches based on the concept of CS have not been thoroughly explored.

\section{Methodology}
\label{sec:methodology}

\subsection{Incorporating Model Uncertainty}

In this work, we propose an adaptive method that considers the estimated uncertainty of the model at time step $t$ to automatically control $k$ and $\alpha$. In other words, our adaptive approach consists in modifying Eq. \eqref{eq:cs} as follows:
\begin{align}
x_t = \underset{v \in V^{(k_t)}}{\arg \max} \Bigg\{(1-\alpha_t) \times \underbrace{p_\theta(v \mid \boldsymbol{x}_{<t})}_{\text{model confidence}} -
\notag
\\
\alpha_t \times \underbrace{\left(\max \{s(h_v, h_{x_j}): 1 \leq j \leq t-1\}\right)}_{\text{degeneration penalty}} \Bigg\}
\label{eq:acs}
\end{align}
where
\begin{align}
    k_t=10*\frac{\exp \left(\delta_t \right)}{\exp (\delta_t)+1}+5
    \label{eq:k_t}
\end{align}

with

\begin{align}
    \delta_t=q*\text{\small arctanh} \left(\frac{H(X)^{(t)}-\text {\small median}(H(X)^{(<t)})}{\text { maximum entropy }}\right)
    \label{eq:delta_t}
\end{align}

and

\begin{align}
    \mathrm{H}(X)^{(t)}=-\sum_{x \in \mathcal{V}} p(x \mid \boldsymbol{x}_{<t}) \ln  p(x \mid \boldsymbol{x}_{<t}).
    \label{eq:entropy}
\end{align}

Once $k$ is selected, a similar procedure is followed to determine $\alpha_t$:

\begin{align}
    \alpha_t=\frac{\exp \left(\delta_{t, k} \right)}{\exp (\delta_{t, k})+1}
    \label{eq:alpha_t}
\end{align}

{\small
\begin{align}
    \delta_{t,k}=q*\text{\small arctanh} \left(\frac{H(X)^{(t, k)}-\text {\small median}(H(X)^{(<t, k)})}{\text { maximum entropy }^{(k)}}\right)
    \label{eq:delta_tk}
\end{align}
}%

In other words, we follow a sequential procedure for $k_t$ and $\alpha_t$ that  involves these steps:

\begin{itemize}
    \item[i)]\textbf{Measuring uncertainty:} Compute the entropy of the output distribution denoted as $H(X)^{(t)}$.
\item[ii)] \textbf{Centering:} Subtract the median entropy of the previous prediction steps.
\item[iii)] \textbf{Scaling:} Divide by the maximum entropy. This step aims to obtain a relative measure, ensuring comparability across different vocabulary sizes.
\item[iv)] \textbf{Computation}: Pass the centered and rescaled entropy term through a sigmoid function, yielding the value of $\alpha_t \in (0, 1)$ - or for the case of $k$ - through a rescaled sigmoid function that yields positive integer values.
\end{itemize}

The scaling term \textit{maximum entropy} refers to the entropy of a uniform distribution over a finite set ${x_1, ..., x_{|\mathcal{V}|}}$, where each token has an equal probability of $\frac{1}{|\mathcal{V}|}$. Consequently, this entropy remains constant over time. For a vocabulary of size $|\mathcal{V}|$, the maximum entropy is given by $\ln(|\mathcal{V}|)$, analogously, the maximum entropy for the distribution of the top-$k$ tokens is given by $\ln(k)$.

Additionally, the parameter $q$ serves as a temperature factor, influencing the range of $k$ and $\alpha$ values at each time step. Adjusting $q$ can either broaden or narrow this range: a lower temperature reduces variability, while a higher value allows for larger changes. This impact is demonstrated in Appendix \ref{a:examples}, Figures \ref{fig:example_wikinews_q1}, \ref{fig:example_wikinews_q8},  \ref{fig:example_wikitext_q1}, \ref{fig:example_wikitext_q8}, \ref{fig:example_book_q1} and \ref{fig:example_book_q8}. However, it is important to note that our evaluation is based on a setup with no temperature (i.e., $q = 1$).

\subsection{Theoretical Motivation}

The general question that might arise is to why $k_t$ and $\alpha_t$ should be chosen adaptively, i.e., why there is no global $\alpha$ nor a global $k$ that is optimal at all time steps.
Taking on a more statistical perspective on the problem offers an explanation: The degeneration penalty in constrastive search can be understood as a regularization term, see also \cite{chen2023fidelity}. More precisely, we have:

\begin{prop}\label{prop-reg}
    The degeneration penalty $\max_j \{s(h_v, h_{x_j})\}$ is a function of the penalty $||h_v - h_{x_j}  ||_2$ in statistical Tikhonov-regularization, if the representations $h$ are normalized.  
\end{prop}
    
The proof can be found in Appendix \ref{a:proofs}. 
Classical statistical regularization aims at preventing overfitting by smoothing out the effect of the training data on the model fit. In CS, the degeneration penalty plays a similar role: It attenuates the effect of the model on the chosen output token. Recall that we do not want to select tokens solely based on the model to prevent repetitive text generation. It is a well-known fact that optimal regularization parameters (corresponding to $k_t$ and $\alpha_t$ here) for many statistical models correspond to the signal-to-noise ratio in the data, see e.g. the work by \citet{kimeldorf1970correspondence,rao3shalabh,hastie2009elements,fahrmeir2022regression}. The intuition is straightforward: For a given signal, the more noise in the data, the higher the optimal regularization parameter should be chosen to prevent overfitting to the latter. This motivates our adaptive approach to CS. Instead of choosing fixed $k$ and $\alpha$ in the beginning, we choose it based on the observed variation of the object on which we want to prevent overfitting. In contrast to statistical modeling, however, this object is the model output, not the training data. We thus choose the optimal $k_t$  and $\alpha_t$ based on the variation of the model output, i.e., its estimated uncertainty measured by $\delta_t$ and $\delta_{(t, k)}$, the standardized Shannon entropy. While several approaches to quantifying uncertainty exist, see \citet{abdar2021review} for an overview, we rely on the classical Shannon entropy since it is computationally efficient and tailored to measuring epistemic (reducible) \textit{predictive} uncertainty, see \cite[section 3.3]{hullermeier2021aleatoric}, as required here.

\begin{figure*}[!ht]
    \centering
    \includegraphics[width=0.9\textwidth]{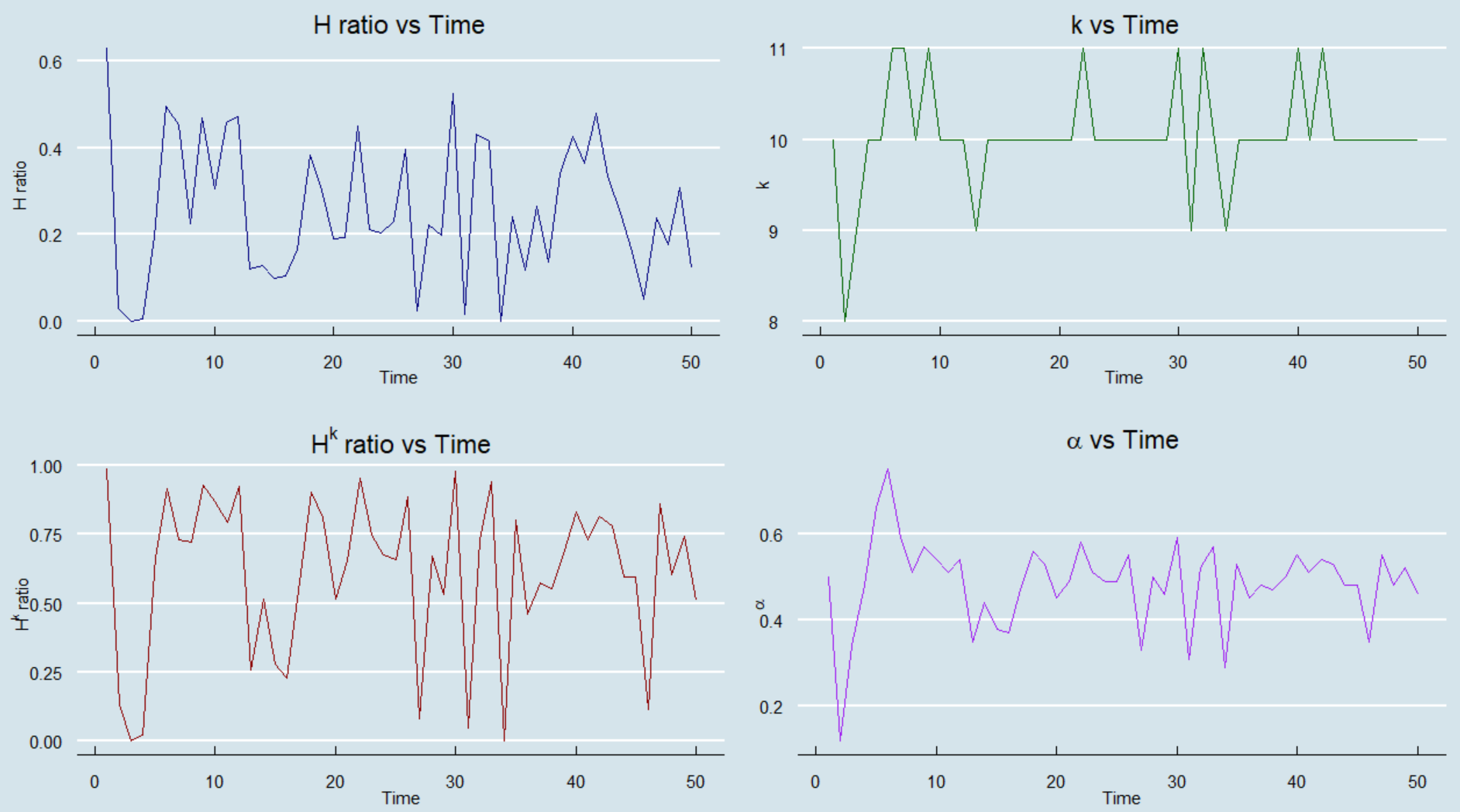}
\caption{Visualization of uncertainty over time, measured by the Shannon entropy of the output distribution (first row, left). It is used to determine the value of $k$ over time (right). The second row illustrates the entropy of the top-$k$ tokens distribution, which is used to compute the value of $\alpha$ (right).}
   
\vspace{0.5em} 
    \begin{minipage}{1\textwidth}
        \small
        \raggedright 
        \textbf{Prompt:} \textit{Knowing that she would be staying in, she started by choosing a pair of fitted, soft, black slippers, the type that barely covered her feet but gave}
  \\        
        \textbf{Generated story:} \textit{her a sense of comfort. As she walked to the dining room, she took a moment to admire the decor and thought about what she would do for dinner. Her family was going to be here for a few days, and she wanted to make the most of the time they had before they left.}
    \end{minipage}
    \label{fig:decoding_behavior} 
\end{figure*}

\begin{table*}[!ht]
\centering
\resizebox{1\textwidth}{!}{
\begin{tabular}{|c|c|c|c|c|c|c|c|c|c|}
\hline \multirow{2}{*}{ Method } & \multicolumn{3}{c|}{ Wikinews } & \multicolumn{3}{c|}{ Wikitext } & \multicolumn{3}{c|}{ Story } \\
 & div.(\%)↑ & MAUVE(\%)↑ & coh.↑ & div.(\%)↑ & MAUVE(\%)↑ & coh.↑ & div.(\%)↑ & MAUVE(\%)↑ & coh.↑ \\
\hline Greedy Search* & 3.55 & 13.96 & \textbf{-0.47} & 1.77 & 4.91 & \textbf{-0.41} & 0.86 & 2.65 & \textbf{-0.34} \\
 Top- $k$ Sampling* & 91.56 & 89.86 & -2.22 & 87.49 & 81.00 & -2.37 & 91.22 & 87.49 & -2.45 \\
 Nucleus Sampling* & 93.54 & 89.45 & -2.61 & 92.16 & 86.54 & -3.03 & 94.50 & 91.47 & -3.02 \\
 Typical Sampling* & 95.37 & 90.97 & -3.26 & 94.82 & 86.07 & -3.71 & 96.29 & 88.58 & -3.68 \\
 CD* & 91.57 & \textbf{92.20} & -2.16 & 88.02 & \textbf{91.46} & -2.19 & 86.41 & \textbf{93.17} & -2.09 \\
 CS ($k = 5, \alpha = 0.6$) & 93.72 & 84.14 & -1.39 & 89.35 & 77.97 & -1.56 & 93.06 & 84.74 & -1.61 \\
  CS ($k = 10, \alpha = 0.6$) & 96.30 & 87.53 & -1.73 & 94.09 & 77.97 & -1.93 & 95.46 & 84.96 & -1.91 \\
 ACS (Ours, $q = 1$) & 95.22 & 79.45 & -1.60 & 92.72 & 78.67 & -1.74 & 93.89 & 80.72 & -1.71 \\
 \hline\hline
Bonus: \textit{DoubleExp} &  \textbf{97.39} &  90.65 & -2.12 & \textbf{96.58} & 84.07 & -2.18 & \textbf{97.37} & 85.66 & -2.16 \\
\hline
\end{tabular}

}
\caption{Automatic evaluation results: 
Numbers marked with * are obtained using the generated texts originally released by \citet{li2023contrastive}, CS results are taken from \citet{su2022empirical}. The highest scores are highlighted in \textbf{bold}.
}
\label{tab:results_multi_datasets}
\end{table*}

\section{Experimental Setup}
\label{sec:exp_setup}
In this section, we describe the metrics, datasets, baseline models, and human evaluation settings.

\subsection{Evaluation Metrics}
\label{sec:metrics}

We follow \citet{su2022empirical} and use three metrics to automatically measure the quality the generations: Diversity, MAUVE, and Coherence.
\paragraph{Diversity.} This metric aggregates $\mathrm{n}$-gram repetition rates: $$\text{DIV}=\prod_{n=2}^4 \frac{\mid \text { unique } \mathrm{n} \text {-grams }\left(\mathrm{x}_{\text {cont }}\right) \mid}{\text { total } \mathrm{n} \text {-grams }\left(\mathrm{x}_{\text {cont }}\right) \mid}.$$ A low diversity score suggests the model suffers from repetition, and a high diversity score means the model-generated text is lexically diverse.

\paragraph{MAUVE.} MAUVE \citep{pillutla2021mauve} score measures the distribution similarity between the set of generated text and the set of gold references.

\paragraph{Coherence.} Proposed by \citet{su2022contrastive}, the coherence metric is defined as the averaged log-likelihood of the generated text conditioned on the prompt as

$$
\operatorname{COH}(\hat{\boldsymbol{x}}, \boldsymbol{x})=\frac{1}{|\hat{\boldsymbol{x}}|} \sum_{i=1}^{|\hat{\boldsymbol{x}}|} \log p_{\mathcal{M}}\left(\hat{\boldsymbol{x}}_i \mid\left[\boldsymbol{x}: \hat{\boldsymbol{x}}_{<i}\right]\right)
$$

where $\boldsymbol{x}$ and $\hat{\boldsymbol{x}}$ are the prompt and the generated text, respectively; [:] is the concatenation operation and $\mathcal{M}$ is the OPT model (2.7B) \cite{zhang2022optopenpretrainedtransformer}.

\paragraph{Human Eval.} In order to evaluate the quality of the generated text, we consider two critical aspects: fluency and coherence. A fluent piece of text is written in grammatical English and has a natural flow (e.g. excluding unnatural repetition or web formatting). A coherent piece of text should stay on topic with the prompt and avoid unnatural topic drift. We provide five native English speakers with 240 competing continuations (A and B) of the same prompt and ask them to rate their coherence and fluency. Definitions and instructions for the rating process are shown in Appendix \ref{a:humeval}, Figure \ref{fig:human_evaluation_form}.

\subsection{Datasets}

Following previous studies, we evaluate our proposed method on three domains for open-ended text generation: news, Wikipedia articles, and stories. For the news domain, we use articles from Wikinews (2000 examples); for the Wikipedia domain, we use the WikiText-103 dataset \citep[1314 examples;][]{merity2016pointer}; and for the story domain, we use the BookCorpus \citep[Project Gutenberg split, 1947 examples;][]{zhu2015aligning}. Each of the examples contains a prompt and a gold reference i.e. human-generated continuation for evaluation purposes. We extract the prompts and decode 256 tokens for the continuations. Finally, we evaluate the generated text based on both the set of metrics (as described in Sec. \ref{sec:metrics}) and human preferences.

\subsection{Baselines}

We compare ACS to widely used decoding methods (including deterministic and stochastic approaches): greedy search, beam search, top-$k$ sampling, nucleus sampling, typical decoding, CD, and CS with constant $\alpha = 0.6$ and two versions of $k$: 5 and 10. We include the latter for a fair comparison: The CS generations with $k = 10$ achieve better automatic evaluation scores than those generated with $k = 5$. Furthermore, they are more comparable to our method, centered around $k = 10$. Further, we include an additional adaptive method: $\textit{DoubleExp}$, which consists on an exponentiation of the argument of the sigmoid function, with the purpose of reaching values of $\alpha$ closer to 0 or 1. Our goal with this method is to exemplify discrepancies between human judgment and MAUVE. The respective human evaluation results are visualized in Table \ref{tab:human_evaluation_double_exp} and its implementation is described in Appendix \ref{a:doubleexp}.

\subsection{Models}

We explore the relationship between model size and the effect of ACS. For this purpose, we use three open-source autoregressive models: gpt2-xl, gpt2-large, and gpt2-medium \citep{radford2019language}. 

\begin{table*}[!ht]
    \centering
\resizebox{1\textwidth}{!}{
\begin{tabular}{|l|c|c|c|c|c|c|}
\hline
\multirow{2}{*}{Dataset} & \multicolumn{3}{c|}{Coherence} & \multicolumn{3}{c|}{Fluency} \\

 & CS is better & CS and DoubleExp are similar & DoubleExp is better & CS is better & CS and DoubleExp are similar & DoubleExp is better \\
\hline
Wikinews & 56\% & 34\% & 10\% & 32\% & 58\% & 10\% \\

Wikitext & 34\% & 46\% & 20\% & 29\% & 63\% & 8\% \\

Story & 49\% & 31\% & 20\% & 32\% & 58\% & 10\% \\
\hline
\hline
All & 48\% & 36\% & 16\% & 28\% & 62\% & 10\% \\
\hline

\end{tabular}
}\\
\vspace{.5em}
\resizebox{1\textwidth}{!}{
\begin{tabular}{|l|c|c|c|c|c|c|}
\hline
\multirow{2}{*}{Dataset} & \multicolumn{3}{c|}{Coherence} & \multicolumn{3}{c|}{Fluency} \\

 & CS is better & CS and ACS are similar & ACS is better & CS is better & CS and ACS are similar & ACS is better \\
\hline
Wikinews & 24\% & 45\% & 31\% & 16\% & 67\% & 18\% \\

Wikitext & 37\% & 28\% & 35\% & 9\% & 74\% & 18\% \\

Story & 41\% & 27\% & 32\% & 21\% & 59\% & 21\% \\
\hline
\hline
All & 31\% & 32\% & 37\% & 17\% & 62\% & 21\% \\
\hline

\end{tabular}
}
    \caption{Human evaluation results for CS vs. \textit{DoubleExp} (top) and CS vs. ACS (bottom) across different datasets.}
    \label{tab:human_evaluation_double_exp}
\end{table*}

\begin{table*}[!ht]
\centering
\resizebox{1\textwidth}{!}{
\begin{tabular}{|c|c|c|c|}
\hline Prompt & \begin{tabular}{l} 
\textit{Butt criticized Donald's controls in certain situations in the game, as well as the difficulty} \\
\textit{of some levels and puzzles. Buchanan also criticized the controls, calling}
\end{tabular} & div. & coh. \\
\hline Beam & \begin{tabular}{l} 
them " too easy " and " too difficult ". He also criticized the game's graphics, stating that the game's \\
graphics were " too simplistic " and "too simplistic ". He also criticized the game's graphics, stating \\
that the game's graphics were " too simplistic" and " too simplistic ". He also criticized the graphics, \\
stating that the game's graphics were " too simplistic " and " too simplistic ". He also criticized the \\
graphics, stating that the game's graphics were " too simplistic " and " too simplistic ". He also \\
criticized the graphics, stating that the game's graphics were"...
\end{tabular} & 0.02 & 0.605  \\
\hline Nucleus & \begin{tabular}{l} 
them " offensively rudimentary ". After the game was released, Buchanan wrote a number of game \\
reviews about his review of the game, primarily summarizing what he thought the reviewers had \\
enjoyed about it. Buchanan was pleased with how consistently these reviews awarded the game. \\
For one, he praised the puzzles, compared them to Nos. 1 and 2 and enjoyed the construction of the \\
artificial intelligence in the game's forms. He found no fault with the puzzles, even in those situations \\
where they seemed repetitive. According to an Xbox Magazine review, " every time...
\end{tabular} & 0.98 & 0.597  \\
\hline CD & \begin{tabular}{l} 
them " unimpressive " and a " nightmare " of an experience to play with players unfamiliar with \\
Tetris. On the other hand, his opinion was shared by other reviewers, and some were critical of the \\
game's technical design for the Wii version of Tetris. In addition, Tintin's review included a quote \\
from Roger Ebert, who said that Tetris was better than the original game due to its simplicity and \\
ease of play. Ebert's comments were included in the game's DVD commentary, released on March \\
22,2010 . It is unclear if any of the video commentary was taken from ...
\end{tabular} & 0.98 & 0.626  \\
\hline CS  & \begin{tabular}{l} 
them " unimpressive " and a " nightmare " of an experience to play with players unfamiliar with \\
Tetris. On the other hand, his opinion was shared by other reviewers, and some were critical of the \\
game's technical design for the Wii version of Tetris. In addition, Tintin's review included a quote \\
from Roger Ebert, who said that Tetris was better than the original game due to its simplicity and \\
ease of play. Ebert's comments were included in the game's DVD commentary, released on March \\
22,2010 . It is unclear if any of the video commentary was taken from ...
\end{tabular} & 0.98 & 0.626  \\
\hline ACS (Ours, $q = 1$) & \begin{tabular}{l} 
them " a pain in the ass to get used to."\\
On the other hand, his opinion was shared by other reviewers, and some were critical of the \\
game's technical design for the Wii version of Tetris. In addition, Tintin's review included a quote \\
from Roger Ebert, who said that Tetris was better than the original game due to its simplicity and \\
ease of play. Ebert's comments were included in the game's DVD commentary, released on March \\
22,2010 . It is unclear if any of the video commentary was taken from ...
\end{tabular} & 0.98 & 0.629  \\

\hline
\end{tabular}

}
\caption{Case Study: Beam search produces degenerative repetitions while nucleus sampling produces text with incoherent semantics w.r.t. the prefix. Contrastive methods exhibit coherent and fluent text.
}
\label{tab:case_study_1}
\end{table*}

\section{Results}
\label{sec:results}


\subsection{Automatic evaluation results}

The automatic evaluation of generated stories, based on diversity, MAUVE, and coherence are presented in Table \ref{tab:results_multi_datasets}. We observe that CS-based approaches tend to foster diversity, while having a slighter loss of coherence, compared to other decoding methods, such as CD and Typical sampling. An additional common trait is that CS-based approaches exhibit lower MAUVE scores, where CD excels across all three datasets. We observe that the method \textit{DoubleExp}, which consists on an exponentiation of the sigmoid arguments, provides a good balance of high diversity and MAUVE, maintaining coherence values that are lower than non-contrastive methods.

\begin{table*}[ht]
\centering
\resizebox{1\textwidth}{!}{
\begin{tabular}{lccccccccc}
\toprule
 & \multicolumn{3}{c}{Wikinews} & \multicolumn{3}{c}{Wikitext} & \multicolumn{3}{c}{Story} \\
\cmidrule(lr){2-4} \cmidrule(lr){5-7} \cmidrule(lr){8-10}
Method & div.(\%)$\uparrow$ & MAUVE(\%)$\uparrow$ & coh.$\uparrow$ & div.(\%)$\uparrow$ & MAUVE(\%)$\uparrow$ & coh.$\uparrow$ & div.(\%)$\uparrow$ & MAUVE(\%)$\uparrow$ & coh.$\uparrow$ \\
\midrule
ACS, q = 1 & 95.22 & 79.45 & -1.6 & 92.72 & 78.67 & -1.74 & 93.89 & 80.72 & -1.71 \\
ACS, q = 2 & 95.03 & 81.66 & -1.57 & 92.69 & 77.48 & -1.71 & 93.38 & 80.85 & -1.67 \\
ACS, q = 4 & 95.75 & 83.41 & -1.76 & 94.02 & 81.56 & -1.87 & 94.97 & 80.14 & -1.82 \\
ACS, q = 8 & 96.92 & 83.10 & -2.02 & 95.23 & 77.79 & -2.08 & 96.02 & 82.71 & -2.04 \\
ACS, q = 15 & 97.46 & 83.03 & -2.24 & 96.39 & 81.66 & -2.25 & 96.66 & 81.44 & -2.23 \\
ACS, q = 20 & 97.78 & 85.01 & -2.32 & 96.55 & 81.61 & -2.33 & 96.66 & 80.37 & -2.26 \\
\bottomrule
\end{tabular}
}
\caption{Ablation results for diversity, MAUVE, and coherence w.r.t. to the adaptiveness enforced by temperature $q$.
}
\label{tab:ablation_q}
\end{table*}

\begin{table}[ht!]
\centering
\resizebox{0.45 \textwidth}{!}{
\begin{tabular}{>{\raggedright}m{3.5cm} >{\centering\arraybackslash}m{3cm} >{\centering\arraybackslash}m{3cm}}
\toprule
\textbf{Method} & \textbf{sec / story$\downarrow$} & \textbf{\# Tokens / sec$\uparrow$} \\
\midrule
CS ($\alpha$ = 0.6, $k$ = 10) & 11.6 & 21.98 \\
ACS ($q$ = 1) & 15.7 & 16.29 \\
ACS ($q$ = 2) & 15.9 & 16.14 \\
ACS ($q$ = 8) & 16.3 & 15.35 \\
\bottomrule
\end{tabular}
}
\caption{Comparison of generation speed for CS and ACS with different temperatures $q \in \{1, 2, 8\}$. Experiments were conducted with a GPU NVIDIA RTX 3090.
}
\label{tab:speed_comparison}
\end{table}

\subsection{Human evaluation}
\label{subsec:human_results}

The human evaluation scores for the CS and ACS methods, as well as for CS and \textit{DoubleExp}, are displayed in Table \ref{tab:human_evaluation_double_exp}. It is worth mentioning that high MAUVE scores do not always align with human judgments. For instance, evaluators consistently show a preference for CS over \textit{DoubleExp} across all datasets. Conversely, ACS is favored in terms of fluency, and AC is preferred for coherence. Nonetheless, when considering all datasets together, there is a slight overall preference for ACS compared to its non-adaptive version.

\subsection{Qualitative examples}

We present qualitative examples to illustrate the distinct characteristics of different decoding strategies. Table \ref{tab:case_study_1} highlights the generated text variations, while Figure \ref{fig:decoding_behavior}
 visualizes the behavior of key parameters such as entropy, $k$, and $\alpha$.

\subsection{Ablation studies}
\label{sec:abl}

To assess the impact of varying levels of adaptiveness, controlled by the temperature parameter, we conduct experiments for $q \in \{1 \text{ (no temperature)}, 2, 4, 8, 15, 20\}$. The results, summarized in Table \ref{tab:ablation_q}, demonstrate the sensitivity of ACS to different values of $q$. As expected, increasing the temperature leads to higher diversity but at the cost of reduced coherence. Moreover, we observe that in both Wikitext and Story, the MAUVE score begins to decline as the generated texts become excessively diverse and erratic.

\subsection{Generation speed}
\label{sec:latency}

We have measured the average generation speed across all three datasets by varying the temperature $q \in \{1, 2, 8\}$ with respect to our baseline (CS with $k = 10$ and $\alpha = 0.6$). A summary is presented in Table \ref{tab:speed_comparison}, showing a decrease of 35\% in the speed for ACS compared to CS with fixed $\alpha$. A supplementary analysis of this for smaller values of $k$ is provided in Appendix \ref{a:lower_k_perf}.

\begin{table*}[ht]
\centering
\resizebox{1\textwidth}{!}{
\begin{tabular}{lccccccccc}
\toprule
\text{Language} & \multicolumn{3}{c}{\text{Contrastive Search}} & \multicolumn{3}{c}{\text{Adaptive Contrastive Search}} & \multicolumn{3}{c}{$\Delta$} \\
\cmidrule(lr){2-4} \cmidrule(lr){5-7} \cmidrule(lr){8-10}
& \text{div.(\%)}↑ & \text{MAUVE(\%)}↑ & \text{coh.}↑ & \text{div.(\%)}↑ & \text{MAUVE(\%)}↑ & \text{coh.} & \text{div.(\%)} & \text{MAUVE(\%)} & \text{coh.} \\
\midrule
\text{Arabic} & 89.55 & 70.53 & -1.51 & 60.71 & 89.94 & -1.23 & -28.84 & 19.41 & 0.28 \\
\text{Bengali} & 72.48 & 89.87 & -1.24 & 85.17 & 96.31 & -1.34 & 12.69 & 6.44 & -0.10 \\
\text{German} & 97.95 & 72.80 & -2.16 & 93.04 & 42.82 & -1.07 & -4.91 & -29.98 & 1.09 \\
\text{French} & 95.74 & 93.21 & -2.27 & 92.49 & 96.41 & -2.08 & -3.25 & 3.20 & 0.19 \\
\text{Hindi} & 98.99 & 95.95 & -1.00 & 98.90 & 92.99 & -1.00 & -0.09 & -2.96 & 0.00 \\
\text{Japanese} & 50.47 & 72.69 & -0.92 & 39.47 & 83.30 & -1.80 & -11.00 & 10.61 & -0.88 \\
\text{Dutch} & 95.47 & 33.57 & -2.96 & 98.03 & 72.32 & -1.30 & 2.56 & 38.75 & 1.66 \\
\text{Chinese} & 91.42 & 93.28 & -2.39 & 82.55 & 92.76 & -2.26 & -8.87 & -0.52 & 0.13 \\
\bottomrule
\end{tabular}
}
\caption{Comparison across different languages. Positive $\Delta$-values indicate better performance of ACS vs CS.
}
\label{tab:multilingual_results}
\end{table*}

\begin{table*}[!ht]
\centering
\resizebox{1\textwidth}{!}{
\begin{tabular}{llccccccccc}
\toprule
Dataset & Model & \multicolumn{3}{c}{Contrastive Search} & \multicolumn{3}{c}{Adaptive Contrastive Search} & \multicolumn{3}{c}{$\Delta$} \\
\cmidrule(lr){3-5} \cmidrule(lr){6-8} \cmidrule(lr){9-11}
 & & div.(\%)↑ & MAUVE(\%)↑ & coh.↑ & div.(\%)↑ & MAUVE(\%)↑ & coh.↑ & div.(\%) & MAUVE(\%) & coh. \\
\midrule
Wikinews & gpt2-xl & 93.72 & 88.14 & -1.39 & 96.92 & 83.10 & -2.02 & 3.20 & -5.04 & -0.63 \\
 & gpt2-large & 93.80 & 78.55 & -1.44 & 96.55 & 78.84 & -2.06 & 2.75 & 0.29 & -0.62 \\
 & gpt2-medium & 3.66 & 12.86 & -0.56 & 49.88 & 20.25 & -6.22 & 46.22 & 7.39 & -5.66 \\
 \hline
Wikitext & gpt2-xl & 89.35 & 77.97 & -1.56 & 95.23 & 77.79 & -2.08 & 5.88 & -0.18 & -0.52 \\
 & gpt2-large & 89.04 & 73.91 & -1.59 & 95.67 & 80.00 & -2.11 & 6.63 & 6.09 & -0.52 \\
 & gpt2-medium & 2.25 & 4.75 & -0.47 & 64.13 & 10.91 & -5.94 & 61.88 & 6.16 & -5.47 \\
 \hline
Story & gpt2-xl & 93.06 & 84.74 & -1.61 & 96.02 & 82.71 & -2.04 & 2.96 & -2.03 & -0.43 \\
 & gpt2-large & 90.63 & 81.16 & -1.56 & 95.82 & 80.42 & -2.05 & 5.19 & -0.74 & -0.49 \\
 & gpt2-medium & 1.22 & 3.08 & -0.40 & 11.86 & 17.19 & -6.13 & 10.64 & 14.11 & -5.73 \\
\bottomrule
\end{tabular}

}
\caption{Comparison of CS ($k = 10, \alpha = 0.6$) and ACS across different datasets and models of varying size.}
\label{tab:model_size_results}
\end{table*}

\subsection{Application to other languages}
\label{sec:multilingual}

We evaluate the performance of our approach across eight additional languages: Arabic, Bengali, German, French, Hindi, Japanese, Dutch, and Chinese. For this, we utilize pre-trained GPT-2-based architectures of various sizes, comparing the Adaptive Contrastive Search (ACS) to standard Contrastive Search (CS). The detailed results are provided in Table \ref{tab:multilingual_results}. The scores vary greatly across languages and metrics, with particularly strong results in Bengali, French, Hindi, and Chinese. Nevertheless, a clear trend emerges: ACS, with the exception of German, consistently achieves comparable or superior MAUVE scores while maintaining a balance between coherence and diversity, outperforming its static counterpart.

\subsection{Effect of varying model sizes}
\label{sec:model_sizes}

We examine the influence of model size on the quality of text generation, focusing on three different sizes: gpt2-xl (1.50B), gpt2-large (0.76B), and gpt2-medium (0.35B). The generated outputs are evaluated across the Wikinews, Wikitext, and Story datasets. As shown in Table \ref{tab:model_size_results}, there are marked differences in performance, particularly with gpt2-medium, where the diversity under CS is substantially diminished across all datasets. We hypothesize that this is linked to the model's isotropy, where only high values of $\alpha$ can foster diversity in the generations. A visual inspection further supports this observation, as the outputs from gpt2-medium tend to be repetitive and degenerate, resembling those produced by greedy or beam search.

\subsection{Findings about MAUVE}
\label{sec:mauve}

Our human evaluation results, as detailed in Table \ref{tab:human_evaluation_double_exp}, corroborate the findings of \citet{su2022empirical}. The discrepancies observed indicate that MAUVE does not consistently align with human preferences. Specifically, the \textit{DoubleExp} method yields higher MAUVE values compared to CS and ACS. However, human evaluators consistently rate \textit{DoubleExp} as less coherent and fluent than the CS approach with $k = 10$ and $\alpha = 0.6$. Moreover, we identified two considerable issues related to varying the truncation length in pairwise sentence comparisons: (1) Inconsistent results that lead to different conclusions regarding the optimal decoding method, and (2) substantial differences in sample sizes, as illustrated in Appendix \ref{a:mauve}, Table \ref{tab:mauve_inconsistencies}.

\subsection{Interpretability}
\label{sec:interpretability}

We conducted additional experiments employing CS with varying values of $k$ and $\alpha$, measuring the same automatic metrics for human-generated text and examining which combinations of hyperparameters align most closely with the gold references. In Appendix \ref{a:interpretability}, our analysis reveals that extreme parameter settings, such as very low values of $\alpha$, yield very low diversity and MAUVE scores but excessively high coherence (even in combination with high values of $k$). Conversely, very high values of $\alpha$ lead to texts that are overly diverse and incoherent (even with moderate values of $k$). A desired balance emerges from these observations: moderate values of $k$ and $\alpha$, such as $k = 10$ and $\alpha = 0.6$, tend to approximate human references by favoring both diversity and coherence. Our experiments show that ACS frequently generates results within this range, at times favoring either higher diversity or greater coherence, as dictated by the uncertainty-guided regularization.

\section{Discussion and Future Work}
\label{sec:disc_future}

In this study, we introduce a novel approach grounded in a CS framework. It is important to note, however, that the quality of generated text extends beyond just model confidence and diversity. Other factors, like informativeness, trustworthiness, and cohesion (as discussed by \citet{de1981introduction}), contribute to the overall quality of the text. Moving forward, our aim is to expand our adaptable method to encompass these traits, thereby improving the evaluation of text quality through a more holistic approach. Furthermore, we would like to explore other criteria for the automatic selection of $k$ and $\alpha$. A path worth exploring could be related to the ratio between generation perplexity and model perplexity, where the algorithm would penalize deviations from the model perplexity through automatic adaptation. Finally, our study has centered on a specific task: Open-ended text generation. However, the potential influence of this adaptive decoding strategy on various tasks and contexts warrants further investigation. We aim to broaden our research to evaluate its efficacy in Machine Translation and Summarization, particularly within the scope of low-resource languages, where our approach may prove beneficial in scenarios where training examples are scarce. Finally, we wish to explore the potential of our approach beyond base models and analyze the effect after SFT and strategic prompting.

\section{Conclusion}
\label{sec:conclusion}

We introduce an uncertainty-guided adaptive method aimed at enhancing the quality of open-ended text generation outputs. Our approach calculates the estimated uncertainty at each time step using Shannon entropy and leverages this measure to dynamically adjust the weighting between model confidence and the degeneration penalty, as proposed by \citet{su2022contrastive} and \citet{su2022empirical}.
Our experiments demonstrate that this method performs well in terms of coherence and diversity, achieving MAUVE scores comparable to existing methods. It receives high ratings from human evaluators, particularly for the fluency of its outputs. This method requires no hyperparameter tuning and utilizes computational resources similar to the non-adaptive CS. Notably, unlike CD, it does not rely on two separate models, making it more versatile and suitable for various tasks. While it introduces some latency, our comprehensive studies comparing this approach to CS highlight its robustness across multiple languages and model sizes. We encourage further research into adaptive methods for open-ended text generation and advocate for the development of new metrics that better align with human judgments to improve the evaluation of decoding strategies.



\section*{Limitations}
\label{sec:limitations}

The proposed approach represents a potential step forward in enhancing text generation quality from language models. However, a key limitation lies in the method’s narrow focus on two main objectives: model confidence and degeneration penalty. While these are critical for evaluating certain aspects of text quality, they do not fully capture the broad spectrum of desirable traits in generated text, such as informativeness, fluency, accuracy, trustworthiness, coherence, and cohesion. By not incorporating these additional factors into the decoding process, ACS may produce text that, while coherent and diverse, lacks depth or fails to effectively communicate complex or specialized information (and facts). Expanding the focus to include these dimensions could significantly improve the robustness and applicability of the method across diverse text generation tasks.
Another limitation worth considering is the architectural choice of our experiments, which focuses on the family of gpt2-models, in particular in its gpt2-xl version. We plan to extend this analysis to more modern architectures, such as Mistral 7B, \citep{jiang2023mistral7b}, Llama2 7B  \citep{touvron2023llama2openfoundation}, Llama 3.1 8B \citep{dubey2024llama3herdmodels}, Deepseek 7B \citep{deepseekai2024deepseekllmscalingopensource}, Qwen2 \citep{yang2024qwen2technicalreport}, Falcon2 \citep{malartic2024falcon211btechnicalreport}.

Additionally, the application of ACS has primarily been explored in the context of text generation tasks such as language modeling and natural language understanding. Its effectiveness on other tasks, such as machine translation, summarization, or multi-modal tasks, remains largely untested. Each of these tasks poses unique challenges and requirements in terms of text diversity, context preservation, and semantic accuracy. Investigating the adaptability and performance of the approach across a broader range of tasks will be essential for its broader generalizability. Lastly, while various measures of local uncertainty have been explored in the context of ACS, such as KL divergence, variance, perplexity, and entropy, there is still much room for exploration and refinement. Further experimentation and analysis are needed to determine the optimal combination of uncertainty metrics that can reliably produce high-quality, open-ended text generation across diverse domains and languages. However, the modularity of ACS in terms of choosing the $\delta$-functions allows for a seamless further exploration of different approaches and applications.

\section*{Ethics Statement}

We affirm that our research adheres to the \href{https://www.aclweb.org/portal/content/acl-code-ethics}{ACL Ethics Policy}. This work involves the use of publicly available datasets and does not include any personally identifiable information. For our human evaluation, we employed third-party evaluators, ensuring a rate of over \$20 per hour. An ethical concern worth mentioning is the use of language models for text generation, which may produce harmful content, either through intentional misuse by users or unintentionally due to the training data or algorithms. We declare that there are no conflicts of interest that could potentially influence the outcomes, interpretations, or conclusions of this research. All funding sources supporting this study are acknowledged in the acknowledgments section. We have diligently documented our methodology, experiments, and results, and commit to sharing our code, data, and other relevant resources to enhance reproducibility and further advancements in the field.

\clearpage

\section*{Acknowledgments}

We wish to express our gratitude to Daniele Pugno and Nicolò Campagnoli for their technical support and visualizations. Matthias Aßenmacher received funding from the Deutsche Forschungsgemeinschaft (DFG, German Research Foundation) as part of BERD@NFDI, under grant number 460037581. Julian Rodemann acknowledges support by the Federal Statistical Office of Germany within the co-operation project "Machine Learning in Official Statistics" as well as by the Bavarian Institute for Digital Transformation (bidt) and the Bavarian Academy of Sciences (BAS) within a graduate scholarship.


\bibliography{custom}

\clearpage

\appendix

\section*{Appendix}

\section{Example Generations}
\label{a:examples}

\begin{figure}[!ht]
    \centering
    \includegraphics[width=0.45\textwidth]{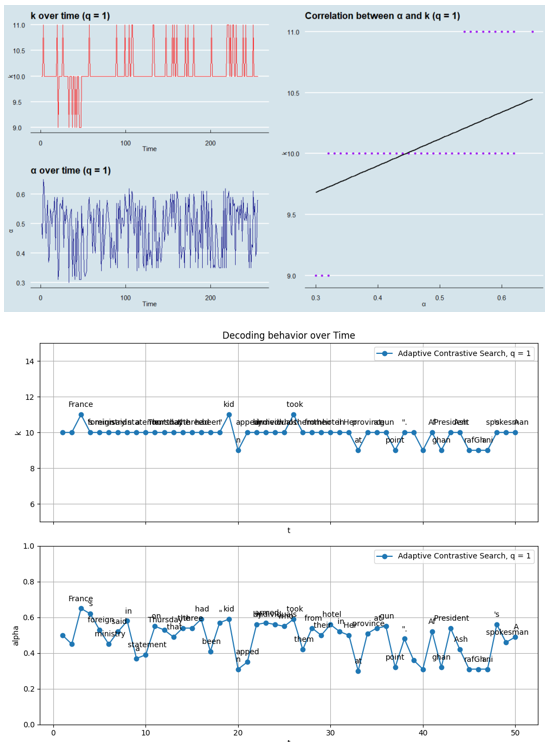}
    \caption{Decoding behavior over time from a Wikinews prompt, $q = 1$.}
    \label{fig:example_wikinews_q1}
\end{figure}
\noindent \textbf{Prompt:} Two French journalists kidnapped on Wednesday north-east of the capital Kabul are believed to be in good health, although the condition of their three Afghan assistants is unknown. \\
        \textbf{Generated story:} France's foreign ministry said in a statement on Thursday that the three had been "kidnapped by armed individuals who took them from their hotel in Herat province at gunpoint". Afghan President Ashraf Ghani's spokesman Aimal Faizi said a helicopter had been sent to pick up the journalists and they were on their way to Bagram airbase in the north-centre of the country. Ghani, who took office in September, has faced criticism for failing to make progress in a war that has claimed more than 2,000 lives since US-led forces overthrew the Taliban in 2001. The United Nations Assistance Mission in Afghanistan (Unama) said this week that security in the country had deteriorated, with attacks on aid workers and foreign military convoys on the rise. Last month, Taliban fighters ambushed a convoy of UN peacekeepers in Kunduz province, killing at least 15 people and wounding more than 30.
(FRANCE 24 with AFP)

\begin{figure}[!ht]
    \centering
    \includegraphics[width=0.45\textwidth]{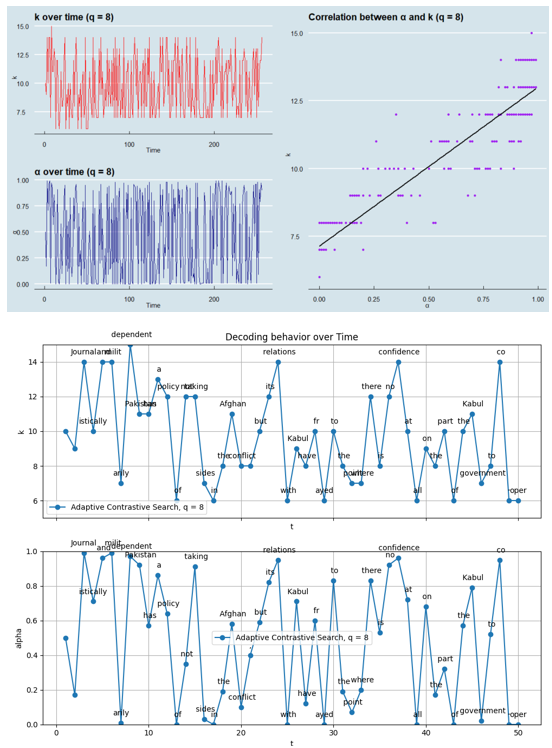}
    \caption{Decoding behavior over time from a Wikinews prompt, $q = 8$.}
    \label{fig:example_wikinews_q8}
\end{figure}
    \vspace{1em} 
\begin{minipage}{0.45\textwidth}
\noindent \textbf{Prompt:} Two French journalists kidnapped on Wednesday north-east of the capital Kabul are believed to be in good health, although the condition of their three Afghan assistants is unknown. \\
        \textbf{Generated story:} Journalistically and militarily dependent Pakistan has a policy of not taking sides in the Afghan conflict, but its relations with Kabul have frayed to the point where there is no confidence at all on the part of the Kabul government to co-operate with Islamabad. Afghan President Hamid Karzai last week cancelled a high-level visit to the United States in protest at Washington's refusal to hand over Bin Laden on humanitarian grounds. Bin Laden escaped from Tora Bora four years ago by flying across the Hindu Kush mountains of Central Asia with help from al-Qaeda-linked guerrillas known as the Haqqani network. Since then his hideout has been under watch by U.S. Special Forces who have set up a command centre close to the Afghanistan-Pakistan border for intelligence, surveillance and counter-terrorism operations.
\end{minipage}
\clearpage

\begin{figure}[ht]
    \centering
    \includegraphics[width=0.45\textwidth]{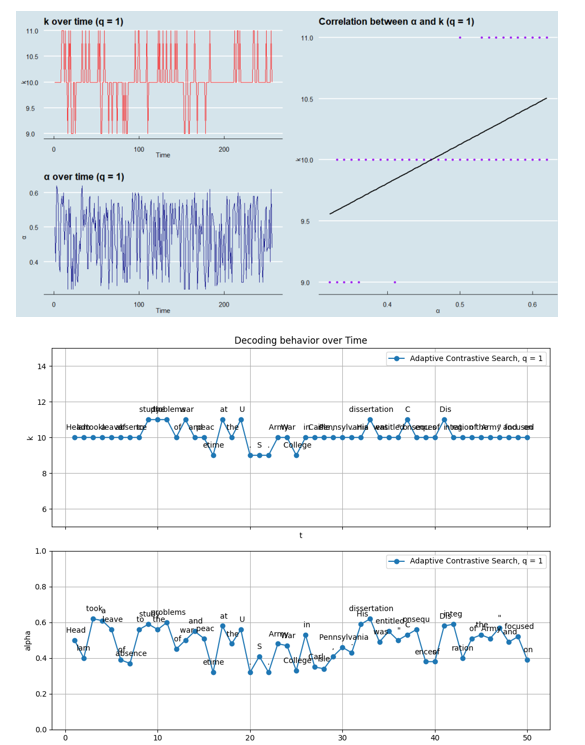}
    \caption{Decoding behavior over time from a Wikitext prompt, $q = 1$.}
    \label{fig:example_wikitext_q1}
\end{figure}
    \vspace{1em} 
\begin{minipage}{0.45\textwidth}
\noindent \textbf{Prompt:} Headlam served as Officer Commanding North @-@ Western Area in 1946, and as Director of Training from 1947 to 1950. In 1950 – 51, \\
        \textbf{Generated story:} Headlam took a leave of absence to study the problems of war and peacetime at the U.S. Army War College in Carlisle, Pennsylvania. His dissertation was entitled "Consequences of Disintegration of the Army" and focused on the effect of demobilization on the quality of officers and NCOs. After the war, Headlam was an instructor at the United States Military Academy at West Point, New York, and served as a military attaché in the Middle East and South Asia. He received the Distinguished Service Medal with Oak Leaf Cluster for his part in the rescue of Lt. Col. (later General) Omar N. Bradley, who was under attack by Japanese at Tarawa, Oahu, in the Solomons (New Hebrides) in 1943. The award was posthumously awarded to him by President Harry S. Truman in recognition of his gallantry in the latter part of World War II. Returning to the United States in the summer of 1951, Headlam began his career as a civilian in the Civil Aeronautics Administration (CAA), where he helped set up the Air Traffic Control System at La Guardia Airport in New York City. The following year, he joined the Federal Aviation
\end{minipage}

\begin{figure}[!ht]
    \centering
    \includegraphics[width=0.45\textwidth]{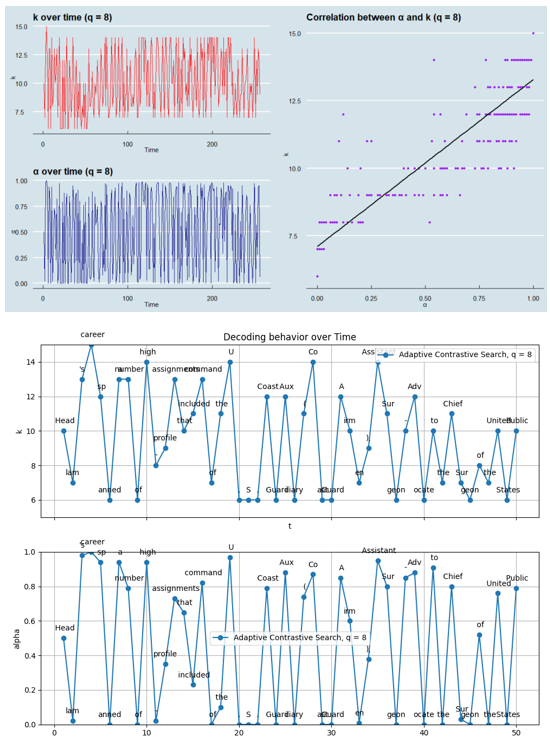}
    \caption{Decoding behavior over time from a Wikitext prompt, $q = 8$.}
    \label{fig:example_wikitext_q8}
\end{figure}
    \vspace{1em} 
\begin{minipage}{0.45\textwidth}
\noindent  \textbf{Prompt:} Headlam served as Officer Commanding North @-@ Western Area in 1946, and as Director of Training from 1947 to 1950. In 1950 – 51, \\        
        \textbf{Generated story:} Headlam's career spanned a number of high-profile assignments that included command of the U.S. Coast Guard Auxiliary (Coast Guard Airmen), Assistant Surgeon-Advocate to the Chief Surgeon of the United States Public Health Service, Special Assistant to the Under Secretary of the Navy for Research and Development, Executive Vice President of the New York-Presbyterian Hospital for 30 years, a Trustee of the Ford Foundation, an Emeritus Visiting Professor at Tufts University in Medford, Mass., and, of course, one of the most prolific authors of UFOlogy to this day. This list goes on and on with every name you can think of – not to mention his work behind the scenes to bring UFOs into the limelight as science rather than superstition. So what can we learn from such a life in and out of the media spotlight? First and foremost, it is easy to lose sight of the human and his/her inner life when looking at the "big picture" of human history. There is nothing wrong with getting paid or doing what you need to do in order to make a living, but that does not mean you are off the hook as a human being.
\end{minipage}

\clearpage

\begin{figure}[!ht]
    \centering
    \includegraphics[width=0.45\textwidth]{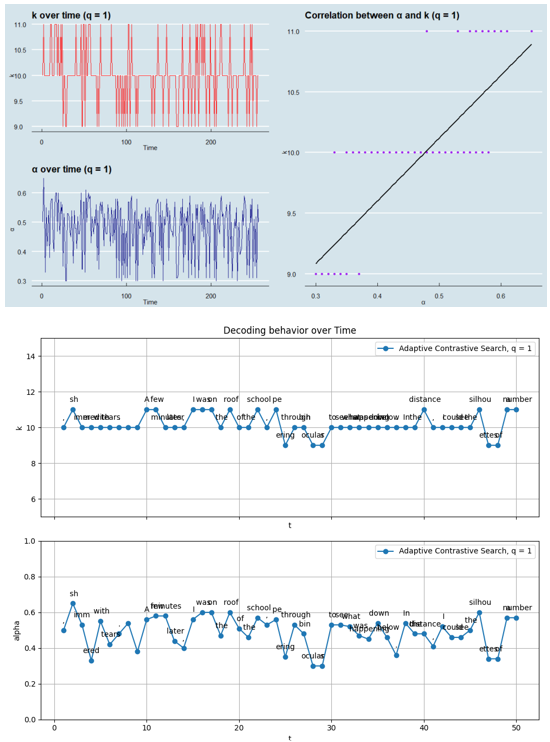}
    \caption{Decoding behavior over time from a Story prompt, $q = 1$.}
    \label{fig:example_book_q1}
\end{figure}
    \vspace{1em} 
    \begin{minipage}{0.45\textwidth}
\noindent  \textbf{Prompt:} The atmosphere dripped with emotion. Even Sola, the 13-year old, caught the current. Her eyes, fixated on the mother-son scene \\
        \textbf{Generated story:} , shimmered with tears. A few minutes later, I was on the roof of the school, peering through binoculars to see what was happening down below. In the distance, I could see the silhouettes of a number of school buses, their windows blackened by fire. It was hard to make out the number of casualties, but by the time I got back to the bus station, there were at least two or three bodies sprawled on the tarmac. The death toll was expected to rise as ambulances ferried the injured to hospitals across the city. Many of the dead were in their 20s and 30s, including a man in his mid-forties who had been driving a minibus when it was hit by a rocket-propelled grenade. His body was strewn across the road, next to a burnt-out car. At the morgue, an elderly man with a moustache and white turban sat in a chair surrounded by coffins. One of the most horrific scenes of the day was the sight of a little girl cradling her mother's mangled body in her arms. She was barely a year old. After the funerals, hundreds of mourners crowded into
    \end{minipage}

\begin{figure}[ht]
    \centering
    \includegraphics[width=0.45\textwidth]{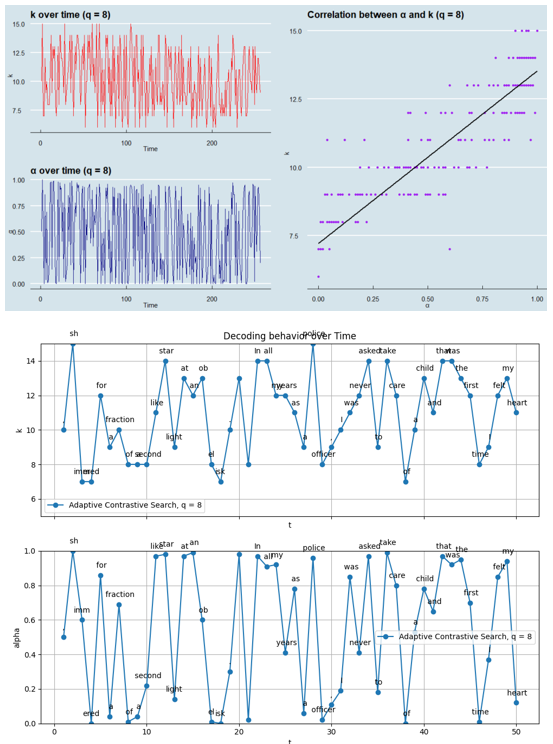}
    \caption{Decoding behavior over time from a Story prompt, $q = 8$.}
    \label{fig:example_book_q8}
\end{figure}
    \vspace{1em} 
    \begin{minipage}{0.45\textwidth}
\noindent  \textbf{Prompt:} The atmosphere dripped with emotion. Even Sola, the 13-year old, caught the current. Her eyes, fixated on the mother-son scene \\        
        \textbf{Generated story:} , shimmered for a fraction of a second like starlight at an obelisk. In all my years as a police officer, I was never asked to take care of a child and that was the first time I felt my heart ache as I walked the beat of the district. There was no mistaking the tears running down my face. After our tour, it was time to eat a celebratory meal that consisted of chicken biryani, roti and samosas. The family seated on the terrace was well-behaved and did not make a peep in our direction. As we chatted amicably, one of the women turned to me and said, "Aunty, what's your job?" I thought it was a good question and tried to find the right answer. "I'm a constable," I said. "That's good," she said without breaking eye contact. "Why are you an constable?" "For two reasons," I told her. "First, it's one of the jobs that requires physical and mental fortitude. The second reason is that in my line of work, every life is precious. You have to make sure that everyone gets a fair
    \end{minipage}

\clearpage

\section{Proofs}
\label{a:proofs}

\paragraph{Proof of Proposition \ref{prop-reg}.}

\begin{proof}
    Per normalization we have for the representations $||h_v||_2 = ||h_v||_2 = 1$. Further, recall that the cosine distance is defined as
    \[
    s(h_v, h_{x_j}) = \frac{h_v^\top h_{x_j}}{||h_v ||_2 \cdot ||h_{x_j} ||_2}
    \]
        We have
    \[
         \begin{array}{ll}
         ||h_v - h_{x_j}  ||_2^2 &= \left( h_v - h_{x_j}\right) \left( h_v - h_{x_j} \right) \\
         &= h_v^\top h_v - 2 h_v^\top h_{x_j} + h_{x_j}^\top h_x \\
         &= 2 - 2h_v^\top h_{x_j} \\
         &= 2 - 2 s(h_v, h_{x_j})
     \end{array}   
    \]
It follows that 
\[
\max_j \{s(h_v, h_{x_j})\} = \max_j \left\{ 2 - \frac{||h_v - h_{x_j}  ||_2^2}{2}\right\},
\]
which was to be shown. $\hfill \square$ 
\end{proof}

\clearpage

\onecolumn

\section{Human Evaluation Form}
\label{a:humeval}
\begin{figure}[ht]
    \centering
    \adjustbox{rotate=90, max width= 1.1\textheight, max totalheight= 1.1\textwidth}{%
        \includegraphics{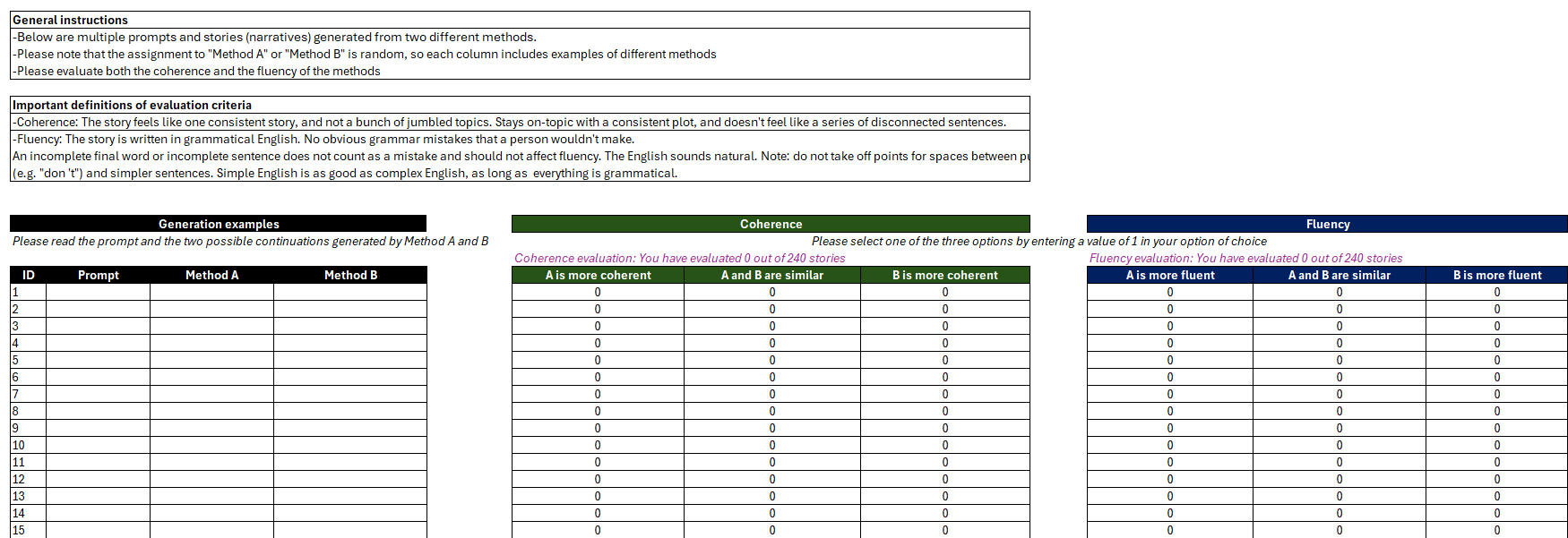}
    }
    \caption{Human evaluation form, including general instructions and definitions for the evaluation criteria.}
    \label{fig:human_evaluation_form}
\end{figure}
\clearpage

\section{Interpretability}
\label{a:interpretability}

\begin{figure*}[!ht]
    \centering
    \includegraphics[width=1.0\textwidth]{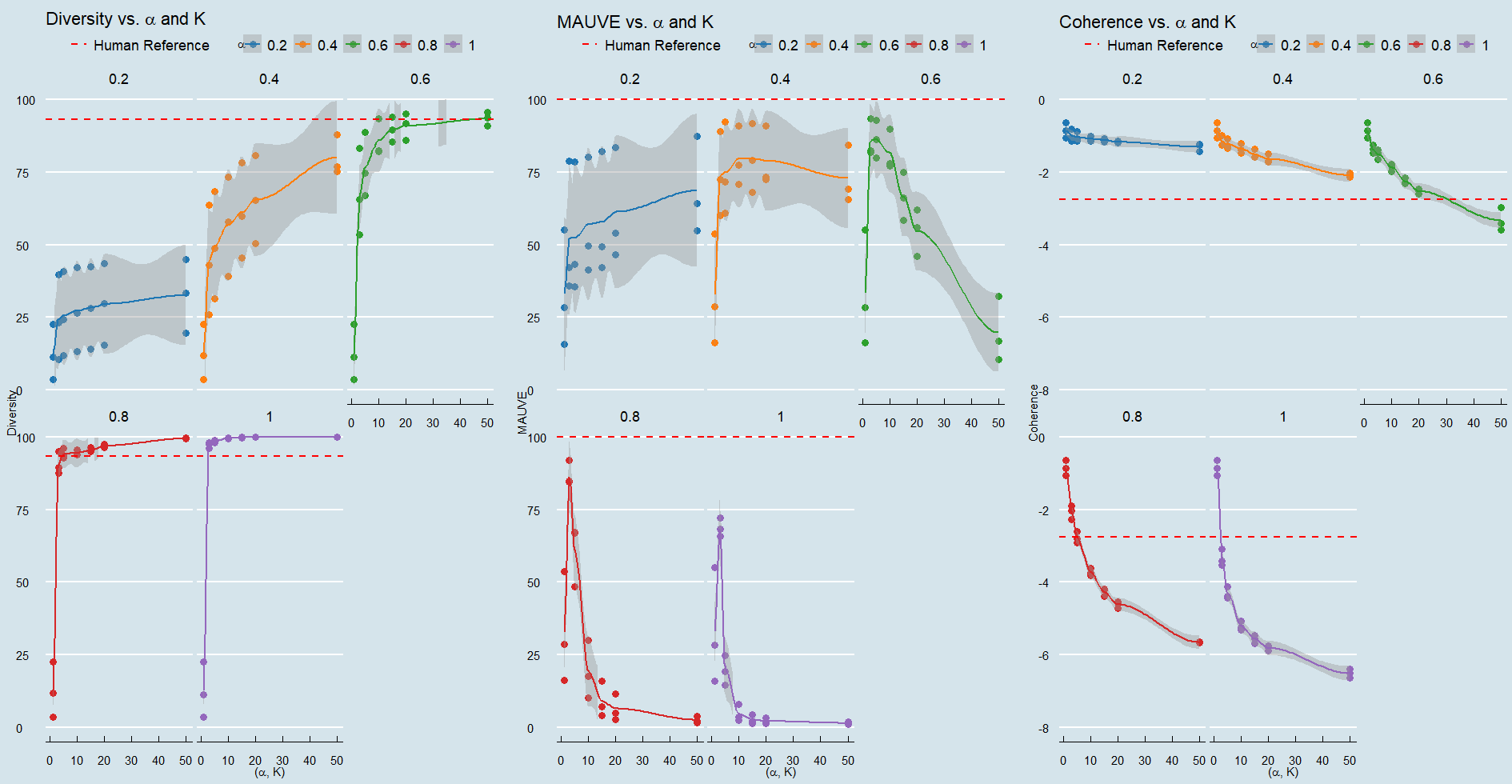}
\caption{To assess the effectiveness of our method, we conducted experiments using Contrastive Search (CS) with varying values of $k \in \{1, 3, 5, 10, 15, 20, 50\}$ and $\alpha \in \{0.2, 0.4, 0.6, 0.8, 1.0\}$. Additionally, we evaluated the diversity, MAUVE, and coherence of human-generated texts from the same datasets, analyzing which hyperparameter combinations most closely align with the gold references. The results indicate that moderate values of $k$ and $\alpha$ tend to produce high-quality generations, closely approximating the performance of human references (dotted red line).}
   
    \label{fig:interpretability} 
\end{figure*}

\section{MAUVE}
\label{a:mauve}

\begin{table*}[ht]
\centering
\resizebox{1\textwidth}{!}{
\begin{tabular}{lccccccc}
\toprule
Dataset & Truncation & \multicolumn{2}{c}{\# Examples} & \multicolumn{3}{c}{MAUVE(\%)↑} & Preferred Method \\
\cmidrule(lr){3-4} \cmidrule(lr){5-7}
 &  & Contrastive Search & Adaptive Contrastive Search & Contrastive Search & Adaptive Contrastive Search &  $\Delta$ &  \\
\midrule
Wikinews & 64 & 1939 & 2000 & 87.42 & 85.79 & -1.63 & Contrastive Search \\
 & 96 & 1920 & 2000 & 81.11 & 88.13 & 7.02 & Adaptive Contrastive Search \\
 & 128 & 1859 & 1977 & 84.14 & 85.39 & 1.25 & Adaptive Contrastive Search \\
 & 160 & 1684 & 1824 & 84.86 & 85.78 & 0.92 & Adaptive Contrastive Search \\
 & 192 & 1447 & 1617 & 85.23 & 87.10 & 1.87 & Adaptive Contrastive Search \\
 \hline
Wikitext & 64 & 1296 & 1314 & 82.78 & 86.83 & 4.05 & Adaptive Contrastive Search \\
 & 96 & 1280 & 1314 & 81.46 & 85.67 & 4.21 & Adaptive Contrastive Search \\
 & 128 & 1250 & 1301 & 77.97 & 79.82 & 1.85 & Adaptive Contrastive Search \\
 & 160 & 845 & 889 & 69.66 & 80.53 & 10.87 & Adaptive Contrastive Search \\
 & 192 & 529 & 564 & 81.50 & 75.45 & -6.05 & Contrastive Search \\
 \hline
Story & 64 & 1907 & 1947 & 84.22 & 87.04 & 2.82 & Adaptive Contrastive Search \\
 & 96 & 1873 & 1947 & 87.82 & 83.66 & -4.16 & Contrastive Search \\
 & 128 & 1657 & 1749 & 84.74 & 85.49 & 0.75 & Adaptive Contrastive Search \\
 & 160 & 863 & 922 & 83.59 & 83.68 & 0.09 & Adaptive Contrastive Search \\
 & 192 & 476 & 518 & 79.43 & 83.38 & 3.95 & Adaptive Contrastive Search \\
\bottomrule
\end{tabular}

}
\caption{MAUVE scores as a function of truncation values, across three different datasets. Positive $\Delta$- values indicate a superior performance of our method. The computations were performed with a gpt2-xl model, for CS we used the reported hyperparameters $k = 5$ and $\alpha = 0.6$.
}
\label{tab:mauve_inconsistencies}
\end{table*}

\nocite{Ando2005,andrew2007scalable,rasooli-tetrault-2015}

\clearpage

\section{DoubleExp Method}
\label{a:doubleexp}

A major concern regarding automatic evaluation metrics in open-ended text generation is their misalignment with human judgment. To illustrate this issue, we introduce a method called \textit{DoubleExp}, which consistently achieves high scores on automatic metrics, yet is systematically rejected by human evaluators, as illustrated in Table \ref{tab:human_evaluation_double_exp}. At each time step $t$, $\alpha$ is dynamically adjusted while maintaining a fixed value of $k = 10$. This approach modifies Eq. \eqref{eq:cs} as follows:
\begin{align}
x_t = \underset{v \in V^{(k)}}{\arg \max} \Bigg\{(1-\alpha_t) \times \underbrace{p_\theta(v \mid \boldsymbol{x}_{<t})}_{\text{model confidence}} -
\notag
\\
\alpha_t \times \underbrace{\left(\max \{s(h_v, h_{x_j}): 1 \leq j \leq t-1\}\right)}_{\text{degeneration penalty}} \Bigg\}
\label{eq:acs}
\end{align}

where

\begin{align}
    \alpha_t=\frac{\exp \left( \text{sgn}(\delta_{t, k}) \cdot \exp(|\delta_{t, k}|) \right)}{\exp \left( \text{sgn}(\delta_{t, k}) \cdot \exp(|\delta_{t, k}|) \right)+1}
    \label{eq:alpha_t_dexp}
\end{align}

with 
{\small
\begin{align}
    \delta_{t,k}= \left(\frac{H(X)^{(t, k)}-\text {\small median}(H(X)^{(<t, k)})}{\text {maximum entropy }^{(k)}}\right)
    \label{eq:delta_dexp}
\end{align}
}

and
\begin{align}
    \mathrm{H}(X)^{(t, k)}=-\sum_{x \in \mathcal{V}^{(k)}} p(x \mid \boldsymbol{x}_{<t}) \ln  p(x \mid \boldsymbol{x}_{<t}).
    \label{eq:entropy_dexp}
\end{align}

\section{Effect for lower values of $k$}
\label{a:lower_k_perf}

In response to concerns about speed limitations, we compared the performance of contrastive search (CS) and our proposed adaptive contrastive search (ACS) across three datasets: Wikitext, Wikinews, and Story. This evaluation focused on key metrics - diversity, MAUVE, and coherence - of the generated texts. Even at lower values of $k$ (specifically, with $k = 5$), ACS demonstrated superior performance, outperforming its static counterpart in 66\% of cases. This improvement was particularly notable in diversity and MAUVE, with only a moderate decrease in coherence. Despite a 32\% reduction in generation speed for ACS compared to standard CS, we do not view this decrease as prohibitive in practical applications. The higher text quality achieved by ACS might compensate for the slower generation time, making it a valuable trade-off for real-world use cases.

\begin{table}[h!] \centering \resizebox{\textwidth}{!}{     \begin{tabular}{lcccccccccccc}
        \toprule
        \multirow{2}{*}{\textbf{Method}} & \multicolumn{3}{c}{\textbf{Wikinews}} & \multicolumn{3}{c}{\textbf{Wikitext}} & \multicolumn{3}{c}{\textbf{Story}} & \multicolumn{3}{c}{\textbf{Average}} \\
        \cmidrule(lr){2-4} \cmidrule(lr){5-7} \cmidrule(lr){8-10} \cmidrule(lr){11-13}
        & \textbf{div.(\%)} $\uparrow$ & \textbf{MAUVE(\%)} $\uparrow$ & \textbf{coh.} $\uparrow$ 
        & \textbf{div.(\%)} $\uparrow$ & \textbf{MAUVE(\%)} $\uparrow$ & \textbf{coh.} $\uparrow$ 
        & \textbf{div.(\%)} $\uparrow$ & \textbf{MAUVE(\%)} $\uparrow$ & \textbf{coh.} $\uparrow$ 
        & \textbf{div.(\%)} $\uparrow$ & \textbf{MAUVE(\%)} $\uparrow$ & \textbf{coh.} $\uparrow$ \\
        \midrule
        \textbf{CS ($\alpha = 0.6, k = 5$)} & 93.72 & 84.14 & -1.39 & 89.35 & 77.97 & -1.56 & 93.06 & 84.74 & -1.61 & 92.04 & 82.28 & -1.52 \\
        \textbf{ACS ($k = 5$)} & 96.16 & 85.39 & -1.71 & 93.28 & 79.82 & -1.79 & 94.53 & 85.49 & -1.74 & 94.66 & 83.57 & -1.75 \\
        \bottomrule
    \end{tabular} } \caption{Comparison of Contrastive Search (CS) and Adaptive Contrastive Search (ACS) across three datasets. Results for diversity, MAUVE, and coherence are reported.} \end{table}

\clearpage

\tableofcontents

\end{document}